%% file: root.tex
\documentclass[letterpaper, 10pt, conference]{ieeeconf}
\IEEEoverridecommandlockouts
\usepackage{balance}
\usepackage{cite}
\usepackage{amsmath,amssymb,amsfonts,bbm,mathtools}
\usepackage{hyperref}
\usepackage{bm}

\usepackage[ruled, linesnumbered]{algorithm2e}
\SetAlCapNameFnt{\footnotesize}
\SetAlCapFnt{\footnotesize}
\usepackage[font=small]{caption} 

\usepackage{graphicx,subcaption}
\usepackage{textcomp}
\usepackage{xcolor}
\usepackage{comment}
\usepackage{multirow}
\let\labelindent\relax
\usepackage{enumitem}
\usepackage{lipsum}
\usepackage{arydshln}
\usepackage{booktabs}
\usepackage{colortbl}
\definecolor{tablegray}{gray}{0.93}

\def\BibTeX{{\rm B\kern-.05em{\sc i\kern-.025em b}\kern-.08em
    T\kern-.1667em\lower.7ex\hbox{E}\kern-.125emX}}

\let\oldnl\nl
\newcommand{\nlnonumber}{\renewcommand{\nl}{\let\nl\oldnl}}

\makeatletter
\let\@oldmaketitle\@maketitle
\renewcommand{\@maketitle}{\@oldmaketitle
    \begin{center}
        \includegraphics[width=0.99\textwidth]{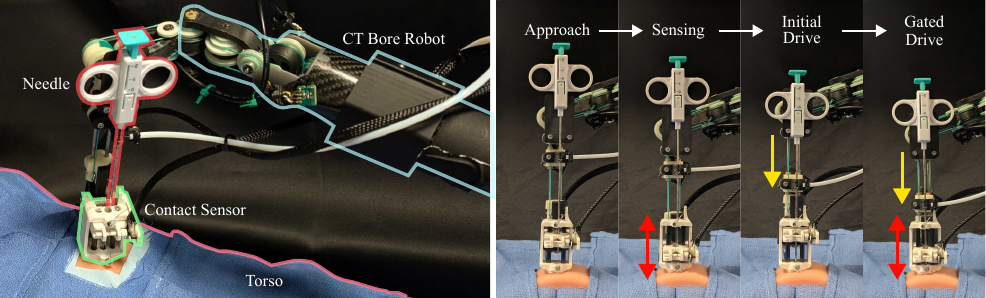}
        \captionof{figure}{We propose a respiratory gated needle insertion system compatible within a CT scanner bore mounted on a CT-guided needle biopsy robot arm. The insertion sequence follows four steps: approaching torso, sense breathing, initial drive, and gated drive. The full sequence is performed while the patient performs quiet breathing, eliminating the need for patient cooperation during breath holds.}
        \label{fig:cover-image}
    \end{center}
    \vspace{-5mm} 
}
\makeatother

\title{\LARGE \bf
Integrated Respiratory Sensing and Gated Control for Autonomous Needle Insertion Via CT Bore Robotic Arm
}
\author{
Emma Huang$^{\dagger,1}$, Moira Bohley$^{\dagger,1}$, Krishna Jamakhandi$^{\dagger\dagger, 2}$, Rebecca Fellbaum$^{\dagger\dagger, 1}$, Sophia Tang$^{3}$\\
Sean Tutton$^{4}$, Albert Hsiao$^{4}$, Alexander Norbash$^{5}$ \\
Changwei Chen$^{1}$, Derek Chen$^{1}$, Ananya Rao$^{1}$, Lizzie Peiros$^{1}$, Michael C. Yip$^{1}$ \IEEEmembership{Senior Member, IEEE}
\thanks{$\dagger$ Equal contribution, $\dagger\dagger$ equal contribution}
\thanks{$^{1}$Department of Electrical and Computer Engineering, University of California San Diego, La Jolla, CA 92093 USA. {\tt\footnotesize\{emh002, dbohley, rfellbaum, chc165, dec004, anr126, epeiros, yip\}@ucsd.edu}}
\thanks{$^{2}$Department of Bioengineering, University of California San Diego, La Jolla, CA 92093 USA. {\tt\footnotesize{kjamakhandi}@ucsd.edu}}
\thanks{$^{3}$Columbia Engineering, Columbia University, New York, NY 10027 USA. {\tt\footnotesize{sht2141}@columbia.edu}}
\thanks{$^{4}$School of Medicine, University of California San Diego, La Jolla, CA 92093 USA. {\tt\footnotesize\{a3hsiao, stutton\}@ucsd.edu}}
\thanks{$^{5}$School of Medicine, University of Missouri-Kansas City, Kansas City, MO 64110 USA. {\tt\footnotesize\{norbash\}@umkc.edu}}%
}

\begin{document}
\maketitle
\begin{abstract}
    Currently, CT-guided lung biopsy is a highly manual procedure requiring patient breath-holding, iterative manual needle insertions, and repeated CT confirmation scans to accurately target a tumor. Automating this process within the CT bore—without relying on manual needle placement or patient breath-holding—has the potential to reduce procedure time, patient discomfort, and radiation exposure. In this work, we present an integrated sensing and control system capable of performing a gated needle insertion within the CT bore. The experiments demonstrate the system's ability to track and predict patient breath cycles, perform autonomous respiratory-gated needle insertion, and detect breath cycle anomalies.
\end{abstract}

\input{sections/1.0_introduction}
\input{sections/3.0_methods}
\input{sections/4.0_results}
\input{sections/5.0_discussion_conclusion}

\bibliographystyle{IEEEtran}
\bibliography{refs}

\end{document}

%% file: sections/1.0_introduction.tex
\section{Introduction}
Lung Carcinoma, commonly known as lung cancer, is the leading cause of cancer-related death worldwide, accounting for 1.8 million deaths per year. Yet, biopsy for diagnosis still faces the challenge of hitting a moving target (a tumor within the lung tissue) and relies on a percutaneous CT-guided needle operated almost entirely by human hands with minimal robotic assistance  \cite{chhikara2023global,saifuddin2021current}. 
In clinical practice, the standard protocol for percutaneous CT-guided needle biopsy involves the clinician walking in and out of the operating room for each needle adjustment, leading to longer procedure times and increased scan radiation exposure for the patient and possibly the clinician or technician.

To address positional errors caused by the patient's natural breathing cycle, physicians have implemented a breath-holding protocol in which the patient holds a stable lung position while the interventional radiologist inserts the needle toward the target. This protocol, while widely adopted, is often unsuccessful for those most likely to need this procedure: older patients with severe pulmonary diseases that make breath-holding difficult \cite{chhikara2023global,saifuddin2021current, carlson2005breathhold, ashraf2017breathhold}. 

Several sensing methods have been considered to address tumor tracking during breath cycles. Contact-based devices such as respiratory belts and bellows provide consistent respiratory signals by tracking changes in chest and abdominal expansion \cite{carlson2005breathhold, locklin2007respiratory}, but they can cause patient discomfort. Optical markers attached to patients can track 3D chest motions through spatially resolved measurements, although they require external cameras and continuous marker visibility~\cite{cala1996chest}. Contactless approaches such as structured light and time-of-flight cameras can also be used to measure chest surface motion without markers, though their effectiveness is influenced by factors including camera positioning, surface visibility, and the physical constraints around the CT scanner~\cite{elshafie2016measuring, schaller2008time, schaerer2012multi}.

Current robotics solutions have made strides in assisting with tumor localization and remove the need for a breath-holding protocol. 
A robotic system for trans-thoracic puncture of pulmonary nodules based on gated respiratory compensation was developed by Li et al.~\cite{li2024gated}. Similarly, Wen et al. ~\cite{wen2026liver} used ultrasound imaging to guide needle insertion with respiratory gating. This method removes the need for patient cooperation in breath-holding protocols, but still requires a surgeon to complete the insertion or manually adjust it after the robotic system completes its measurements. 

Furthermore, Schreiber et al.~\cite{schreiber2019open, schreiber2022crane, schreiber2024crane} and Zhang et al.\cite{zhang2025ct} demonstrated a robotic arm capable of approaching and driving a needle insertion mechanism within the CT bore, eliminating the need to remove the patient from the CT bore for needle insertion. While this system showed potential to reduce operation time and radiation exposure by working directly within the CT bore, it lacks a precise mechanism to compensate for chest movement caused by the patient's natural breathing cycle, limiting its potential to safely perform biopsy procedures. 

In this work, we present an integrated sensing and control system capable of gated needle insertion within the CT bore. This system was developed to integrate into a CT-guided needle biopsy robot~\cite{schreiber2024crane}, addressing current gaps in robot-assisted needle biopsy procedures. 
Our contributions include:
\begin{itemize}
    \item A model for quiet breathing;
    \item An integrated sensing and controls mechanism capable of working within the CT bore;
    \item Autonomous four stage linear needle insertion procedure with gated controls;
    \item Anomaly detection during autonomous procedure.
\end{itemize}
Our experimental results demonstrate successful needle insertion while satisfying respiratory gating constraints derived from sensed breathing motions.

%% file: sections/3.0_methods.tex
\section{Method}
\subsection{Respiratory Motion Characterization} 

\begin{table}[t]
\vspace{2mm}
\centering
\footnotesize
\setlength{\tabcolsep}{3pt}
\renewcommand{\arraystretch}{0.95}
\setlength{\aboverulesep}{1pt}
\setlength{\belowrulesep}{1pt}
\caption{Breathing characteristics and sensing design targets.
Measurements show mean $\pm$ SD and observed range.}
\label{tab:breathing_summary}
\begin{tabular}{@{}lcc@{}}
    \toprule
    \textbf{Parameter} & \textbf{Value} & \textbf{Observed range} \\
    \midrule
    \multicolumn{3}{@{}l}{\textbf{Literature reference values}} \\
    Breath duration (s)
        & 4.335 & -- \\
    Displacement amplitude (mm)
        & 4.33 & -- \\
    Inhalation/exhalation ratio
        & 0.64 & -- \\
    \midrule
    \multicolumn{3}{@{}l}{\textbf{Experimental measurements}} \\
    Breath duration (s)
        & $4.03 \pm 0.97$ & 2.41--6.44 \\
    Displacement amplitude (mm)
        & $4.98 \pm 0.62$ & 3.94--5.72 \\
    Inhalation/exhalation ratio
        & $0.81 \pm 0.12$ & 0.51--1.08 \\
    \midrule
    \multicolumn{3}{@{}l}{\textbf{Sensing design targets}} \\
    Design amplitude range (mm)
        & 3.74--6.22 & -- \\
    Spatial resolution target (mm)
        & 0.37 & -- \\
    Maximum response time (ms)
        & 5 & -- \\
    \bottomrule
    \vspace{-10mm}
\end{tabular}

\end{table}

The sensing system design depends on the timing, amplitude, and direction of chest motion during breathing.
Table~\ref{tab:breathing_summary} summarizes reference measurements from healthy individuals~\cite{dissanayake2026detection,de1997chest,dellaca2015correlated}.
Normal breathing follows an approximately periodic but asymmetric pattern, with expiration generally lasting longer than inspiration~\cite{dellaca2015correlated}.
The chest moves in three dimensions during breathing, primarily along the dorsoventral direction, and the displacement varies across anatomical locations~\cite{hazari2021conceptual}.

\begin{figure}[t]
    \vspace{2mm}
    \centering
    \includegraphics[width=\linewidth]
        {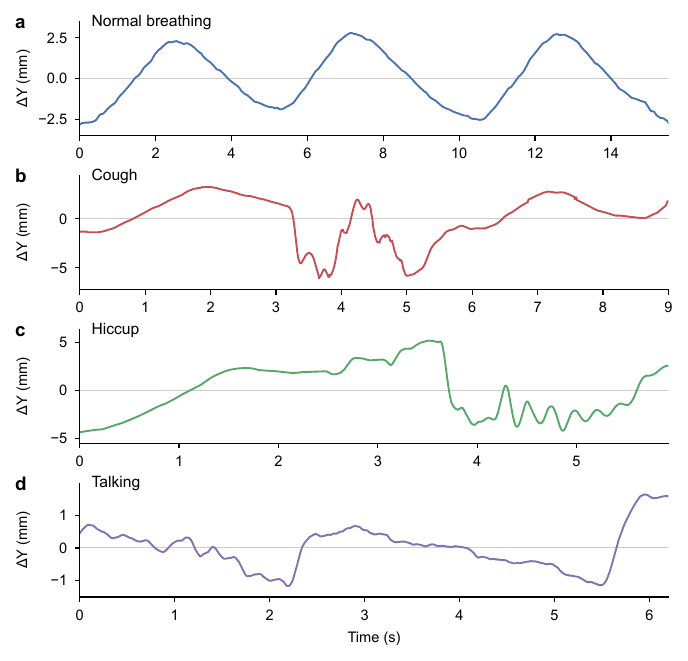}
    \caption{Representative dorsoventral chest displacement
    trajectories recorded with OptiTrack during
    (a) normal breathing, (b) coughing, (c) hiccups,
    and (d) talking.}
    \label{fig:breathing_modes}
    \vspace{-5mm}
\end{figure}

Chest motion from eight volunteers was recorded to compare breathing characteristics with the reference values and obtain trajectories for testing the sensing and control system.
Nine OptiTrack markers were placed in a $3\times3$ grid on each volunteer's chest, with the upper row just below the collarbones and the lower row just above the lower edge of the ribcage.
Each volunteer breathed naturally during a 10-minute recording, and the dorsoventral component of the recorded chest motion was extracted.
Fig.~\ref{fig:breathing_modes} shows approximately periodic chest motion during normal breathing, abrupt changes during coughing and hiccups, and smaller, irregular fluctuations during talking.
Table~\ref{tab:breathing_summary} reports the measured breath duration, displacement amplitude, and inhalation-to-exhalation duration ratio.
The measured mean amplitude of 4.98 mm is comparable in magnitude to the reference sternum displacement of 4.33 mm.

This system must sense and react to typical and abnormal breath cycles.
We use these measurements to define the sensing design targets in Table~\ref{tab:breathing_summary}.
The measured breathing amplitude has a mean of 4.98 mm and a standard deviation of 0.62 mm.
A lower reference amplitude is defined as the mean minus two standard
deviations, yielding 3.74 mm.
As a design criterion, the sensor should distinguish approximately 10 spatial increments across this reference displacement, giving a target spatial resolution of 0.37 mm.
For safety, signals outside the distribution must be detected quickly to prevent unintentional insertion. Examples include hiccups and sneezes, which, while rare, could occur during operation. 
These events are signified by rapid thoracic motion within 35 ms and 20 ms, respectively. 
The desired reaction time for these types of events would be 5 ms.

    \begin{figure*}
    \vspace{2mm}
        \centering
        \includegraphics[trim= 0 22.8cm 0 .8cm, clip, width=0.99\linewidth]{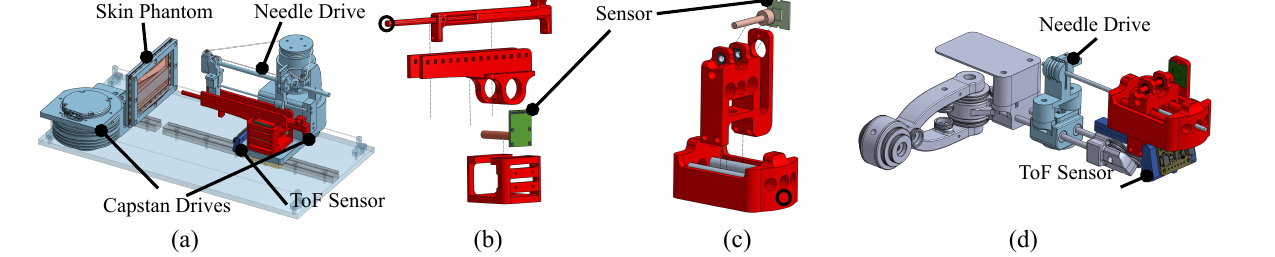}
        \caption{Assembly (a) shows the testbed used to emulate CT bore robot motion and to run tests within a controlled system. The design features a linear rail with two carriages driven by motorized capstan drives. Attached to the breathing model drive is a replaceable skin phantom that can be punctured. The needle insertion mechanism is used to control and drive the needle motion and has a clamp-based mechanism to allow the needle to be fixed and also exchanged after use. Assemblies (b) and (c) show a zoomed view of the contact sensors for the testbed and CT bore robot respectively, each of which read through a Hall effect sensor. Assembly (d) shows the assembly on the CT bore robot. The open circles in (b) and (c) indicate the point of contact on the contact sensor.}
        \label{fig:Bench_top}
        \vspace{-5mm}
    \end{figure*}

\subsection{Mechanical Design}

In order to accurately render the breathing profile displacement, a bench-top breathing mechanism was created as a simplified torso skin phantom which provides a platform to run breathing profiles both synthetic and real. A version of the needle driver platform from a CT-guided needle biopsy robot~\cite{schreiber2024crane}, which will be referred to as the CT bore robot, was created and featured the sensing and actuation to isolate the functional components being designed and assess their capabilities. Finally, a separate CT-compatible version was also created to integrate with the CT bore robot.

\subsubsection{Bench Top Breathing Mechanism}

The testbed takes in a motion profile and outputs a single-degree-of-freedom translational output to mimic the motion of the outer layer of skin around the chest.

The \textit{Breathing Mechanism} is a skin phantom holder on a capstan drive which simulates the motion of breathing, shown in Fig.~\ref{fig:Bench_top}(a). Each skin phantom is made with 00-20 silicone and secured over a 0.625" 60A durometer silicone rubber layer to replicate the compliance of the epidermis and dermis respectively with a 5 by 15 mm rectangular cube secured behind to model the stiffness the rib cage provides beneath the skin~\cite{vandemeulebroucke2011chest}. The phantom is mounted to a carriage on a linear rail which is driven with a CubeMars AK60-V3 motor using a capstan drive mechanism. The capstan drive can reproduce the 95th percentile breathing displacement (6.22 mm) and meet the velocity requirements for travel within the average breath duration (4.335s) seen in Table~\ref{tab:breathing_summary}.

\subsubsection{Bench Top Needle Insertion Mechanism}

The \textit{Needle Mechanism} simulates the similar mechanism on the CT bore robot and the one degree of freedom (DOF) of the wristed end. This is shown in Fig.~\ref{fig:Bench_top}(a). The carriage driver (1 DOF wrist) uses a capstan drive mechanism powered by a CubeMars GL60II Gimbal Motor, while the needle insertion mechanism powered by a CubeMars GL40II Gimbal Motor, providing a representation of the clutch-unclutch needle insertion mechanism on the CT bore robot.

\subsubsection{Bench Top Sensor Mechanism}\label{subsubsec:sensor-mech}
The carriage carries two sensing systems: a VL53L4CD ToF (Time of Flight) sensor and an AS5048B Hall effect sensor detailed in Fig.~\ref{fig:Bench_top}(a,b). The ToF sensor is used to detect distance while approaching to make initial contact with the phantom. It is placed on the front of the carriage with the sensor focal point angled to match with the needle tip and contact sensor. Once in contact with the surface the contact sensor is designed to measure real-time chest displacement. The contact sensor drives a spring-tensioned capstan drive to convert linear displacement to rotation, which is read by rotating a magnet in front of the hall effect sensor shown in green in Fig.~\ref{fig:Bench_top}(b).

\subsubsection{CT Integrated Needle Mechanism and Sensor Mechanism}
The CT-integrated needle sensing system is a compact and optimized version of the bench-top needle mechanism. The CT-integrated contact sensor mechanism, shown in Fig.~\ref{fig:Bench_top}(c), is similarly compacted for CT-integration and designed to minimize metal components for compatibility with a CT scanner. Similarly, the ToF sensor is mounted in the front and angled to measure near the needle tip and the edge of the contact sensor.

Rather than a capstan drive, the needle driver on the CT uses a ``clutch-unclutch" mechanism shown in Fig.~\ref{fig:Bench_top}(d). This action is achieved using a Dyneema string within a Bowden cable, which, when tensioned, tightens a collar around the needle. It can be activated when desired and left inactive when the needle needs to move freely, depending on the stage of the breathing cycle. While a direct-drive motor could presumably provide cleaner dynamics and direct motion measurements, CT imaging is negatively impacted by dense metals, which can introduce artifacts and distort the scans.

\subsection{Sensors} \label{subsec:sensors}
The choice of which sensor reading to use is selected based on a \textit{contact threshold} set on the contact sensor. The measurement and noise are taken individually per sensor:
\vspace{-2mm}
\begin{gather}
    x_{meas} =
    \begin{cases}
        x_{tof}, & \text{no contact} \\
        x_{contact}, & \text{contact}
    \end{cases}
    \\
    \sigma_{meas}^2 =
    \begin{cases}
        \sigma_{tof}^2, & \text{no contact} \\
        \sigma_{contact}^2, & \text{contact}
    \end{cases}
\end{gather}
The contact sensor is initialized at $x_{zero}$ while fully extended. Contact is detected when $(x_{contact} - x_{zero}) \ge ( d_{thresh} + k \cdot \sigma_{contact})$.
During no contact, the ToF sensor serves as the sole measurement source across $[d_{thresh}, 1200mm]$. Once contact is detected, the contact sensor becomes the sole measurement source. A handoff at $d_{thresh}$ is chosen over sensor fusion due to the difference in range and accuracy of the two sensors. 
Both sensors are sampled at their own native rate and updated asynchronously.
The measurement rate is then capped at $\tau_{rate}$ during the control loop.

\subsection{Controls} \label{subsec:controls}
\subsubsection{Delay characterization}
Respiratory-gated needle insertion requires the controller to act precisely on the target's position rather than a delayed sensor reading of that position. Because the target is continuously in motion, any lag between measurement and action introduces positional error proportional to the target's velocity during that lag. Therefore, a forecast horizon should be defined that the system can foresee to account for these delays, factoring all delay sources as,
\vspace{-2mm}
\begin{equation}
\tau_{horizon} = \tau_s + \tau_c + \tau_{cl}(\omega_r) + T_{ins},
\label{eq:horizon}
\end{equation}

where $\tau_s$ is the lag from the sensor to the target's true position. This includes sensor latency but also the contact mechanical compliance. $\tau_c$ is computation delay for the estimation and prediction latency. $\omega_r$ is the rate of the breathing cycle and $\tau_{cl}(\omega_r)$ is the tracking lag of the needle-position servo, which is defined by how the lead compensated closed-loop controller is characterized. $T_{ins}$ is the flight time of the needle to its designated position.

\subsubsection{Signal construction}

The respiratory model as well as the collected data characterize the breath cycle as periodic with asymmetric inhale and exhale patterns. Real human breathing causes small shifts in the frequency of the breathing (Table~\ref{tab:breathing_summary}). 
The signal is modeled as a finite harmonic sum with a slowly-varying phase:
\vspace{-2mm}
\begin{equation}
x(t) = a_0 + \sum_{k=1}^{K} A_k \sin\!\big(k\,\theta(t) + \phi_k\big), \qquad \dot\theta(t) = \omega_r(t),
\label{eq:harmonic_model}
\end{equation}

where $a_0$ is the amplitude, $A_k$ is the harmonic amplitude, $\phi_k$ is the harmonic phase, and $K$ is the harmonic order.

$\omega_r(t)$ drifts over time due to the breath to breath variability of a given patient. In order to construct the signal for online tracking, a calibration window must be recorded. The longer this window is, the more accurate the initial estimate may be, but due to the unrealistic nature of having a patient wait extended periods of time, a $60$ second window that captures approximately 15 breath cycles is used. From this a Fast Fourier transform is taken to get an initial estimate $\hat{\omega_r}$\cite{dissanayake2026breathing}. 

An Ordinary Least Squares (OLS) regression is used to initialize the harmonic amplitudes $A_k$ and phases $\phi_k$ by converting Eq.~\ref{eq:harmonic_model} into a matrix form $y = X\beta + \varepsilon$, where the parameters are put into $\beta = [a_0, \alpha_1, \beta_1, \ldots, \alpha_{K}, \beta_{K}]^T$.
\begin{align}
X_{n,1} &= 1, \qquad
X_{n,2k} = \sin(k\hat\omega_rt_n) \\
X_{n,2k+1} &= \cos(k\hat\omega_rt_n), \quad k=1,\ldots,K \\
\hat\beta &= \left(X^TX\right)^{-1}X^Ty \nonumber \\ 
&= \begin{bmatrix}\hat{a}_0, \hat{\alpha}_1, \hat{\beta}_1, \dots, \hat{\alpha}_{K}, \hat{\beta}_{K}\end{bmatrix}^T \label{eq:ols_estimate}\\
\mathrm{Cov} &= \sigma^2 \left(X^TX\right)^{-1} \label{eq:ols_cov}
\end{align}
By evaluating $\hat\beta$, we can extract the amplitude and phase for the $k$-th element as follows.
\begin{equation}
A_k = \sqrt{\hat\alpha_k^2 + \hat\beta_k^2}, \qquad \phi_k = \operatorname{atan2}(\hat\beta_k, \hat\alpha_k)
\label{eq:amp_phase}
\end{equation}

$K$ is chosen to recover 95\% of the total energy.

Real respiratory rate and depth drift breath to breath, motivating an online, recursive estimator. For a discrete-time nonlinear system with state $\mathbf{s}$, process model $f$, and measurement model $h$, an extended Kalman filter maintains a running estimate of the state and its covariance, alternating between predicting how that estimate evolves under $f$ and correcting it against the next measurement. Because $h$ is nonlinear, each correction linearizes it about the current estimate rather than solving one fixed linear system for every step.
\begin{align}
\hat{\mathbf{s}}_{k|k-1} &= f\!\left(\hat{\mathbf{s}}_{k-1|k-1}\right), \label{eq:ekf_predict_state}\\
P_{k|k-1} &= F_k\, P_{k-1|k-1}\, F_k^T + Q, \label{eq:ekf_predict_cov}\\
\nu_k &= z_k - h\!\left(\hat{\mathbf{s}}_{k|k-1}\right), \label{eq:ekf_innovation}\\
S_k &= H_k\, P_{k|k-1}\, H_k^T + R, \label{eq:ekf_innov_cov}\\
\mathbf{K}_k &= P_{k|k-1}\, H_k^T\, S_k^{-1}, \label{eq:ekf_gain}\\
\hat{\mathbf{s}}_{k|k} &= \hat{\mathbf{s}}_{k|k-1} + \mathbf{K}_k \nu_k, \label{eq:ekf_update_state}\\
P_{k|k} &= \left(I - \mathbf{K}_k H_k\right) P_{k|k-1} \left(I - \mathbf{K}_k H_k\right)^T + \mathbf{K}_k R \mathbf{K}_k^T. \label{eq:ekf_update_cov}
\end{align}
\begin{figure*}[t]
    \vspace{2mm}
    \centering
    \includegraphics[width=.99\linewidth, trim={0 665 0 0}, clip]{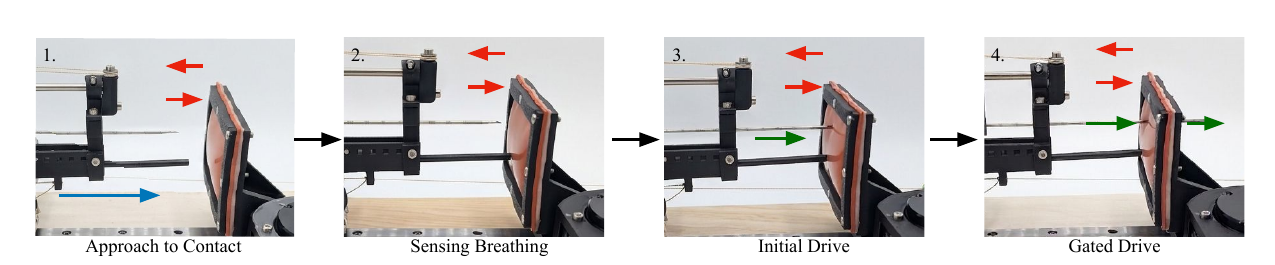}
    \caption{The gated needle insertion process involves four steps: (1) The robot base and wrist bring the needle to a position suitable for insertion. (2) The contact sensor is brought within range of the torso to maintain contact and sense breathing. (3) Once the breathing motion is captured, the needle is driven into the skin. (4) The needle is allowed to passively move with the breathing motion, applying a gated insertion when the suitable breathing stage is detected.}
    \label{fig: state_mach}
    \vspace{-5mm}
\end{figure*}
where $F_k = \partial f/\partial \mathbf{s}\big|_{\hat{\mathbf{s}}_{k-1|k-1}}$ and $H_k = \partial h/\partial \mathbf{s}\big|_{\hat{\mathbf{s}}_{k|k-1}}$ are the Jacobians linearizing $f$ and $h$ about the current estimate. $Q$ is the process-noise covariance, $R$ the measurement-noise covariance, $P$ the state-error covariance, $v_k$ the innovation, $S_k$ the innovation covariance, and $\mathbf{K}_k$ the Kalman gain. Our sensor is scalar, so $H_k$ is a $1\times n$ row vector, $R$ and $S_k$ are scalars, and Eq.~\ref{eq:ekf_gain} reduces to a division rather than a matrix inversion.

With harmonic order $K$ being fixed, the state has dimension $n = 2K + 3$, and is,
\vspace{-2mm}
\begin{equation}
\mathbf{s} = \begin{bmatrix} a_0 & A_1 & \phi_1 & \cdots & A_K & \phi_K & \theta & \omega_r \end{bmatrix}^T.
\label{eq:state_vector}
\end{equation}

The absence of a predictive model for breath rate drift is accounted for by the process-noise covariance $Q$. The state evolves over sample period $T_s$ as:
\vspace{-2mm}
\begin{align}
\theta_k &= \theta_{k-1} + \omega_{r,k-1}\, T_s, & s_k(i) &= s_{k-1}(i), \quad i \neq \theta. \label{eq:process_rw}
\end{align}

The scalar measurement function is the harmonic sum evaluated at the current phase estimate:
\vspace{-2mm}
\begin{equation}
h(\mathbf{s}) = a_0 + \sum_{k=1}^{K} A_k \sin(k\theta + \phi_k).
\label{eq:meas_model}
\end{equation}

The row-vector Jacobian $H_k$ is analytic. Since $\partial h/\partial a_0 = 1$ and $\partial h/\partial \omega_r = 0$, the remaining components are:
\vspace{-2mm}
\begin{align}
\frac{\partial h}{\partial A_k} &= \sin(k\theta + \phi_k), \quad \frac{\partial h}{\partial \phi_k} = A_k \cos(k\theta + \phi_k), \label{eq:jac_Ak_phik}\\
\frac{\partial h}{\partial \theta} &= \sum_{k=1}^{K} k\, A_k \cos(k\theta + \phi_k), \label{eq:jac_theta}
\end{align}

With $\mathbf{s}$, $F_k$, $H_k$, $Q$, $R$, and $P$ all now defined above (Eqs.~\ref{eq:ekf_predict_cov}--\ref{eq:ekf_update_cov}), their initial values are determined from the calibration recording as follows:

\begin{itemize}
\item \textbf{$\hat{\mathbf{s}}_0$} --- read directly from the batch regression ($a_0, A_k, \phi_k, \hat\omega_r$), with $\theta$ initialized at the fitted phase value at the end of the calibration window.
\item \textbf{$P_0$} --- the $a_0/A_k/\phi_k$ block follows from the regression's coefficient covariance, Eq.~\ref{eq:ols_cov}, converted to amplitude/phase form via a first-order (delta-method) Jacobian; $\omega_r$'s initial variance is set from the frequency search's resolution, $2\pi/(N T_s)$.
\item \textbf{$R$} --- estimated from sensor variance during a still/breath-hold segment where available.
\item \textbf{$Q$} --- estimated empirically by refitting each breath in the calibration recording separately, computing the variance of each state across those per-breath fits, and rescaling from breath-timescale to sample-timescale:
\end{itemize}

\begin{equation}
Q_{ii} \approx \mathrm{Var}_{\text{breath-to-breath}}(i) \cdot \tfrac{T_s}{T_{breath}},
\label{eq:Q_construction}
\end{equation}
where $Q$ is a diagonal matrix and a scalar multiplier $q_{scale}$ can then be applied across this diagonal to tune the filter's performance. $T_{breath}$ is the duration of one breath cycle.

Two states in $Q$ are exceptions to Eq.~\ref{eq:Q_construction}. $\theta$ has no independent per-breath estimate of its own. Because it acts as the running phase, it inherits $\phi_1$'s variance directly. $\omega_r$ cannot be estimated the same way. The per-breath refits above hold $\hat\omega_r$ fixed by construction, so there is no breath-to-breath spread in it to measure. Its entry instead comes from re-estimating the frequency independently over several overlapping multi-breath windows spanning the calibration recording and taking the variance across those windows. A single breath is too short a window to resolve frequency precisely enough for its scatter to be meaningful. When the calibration recording is too short to form at least two such windows, this falls back to the same resolution-based value used to seed $\omega_r$ in $P_0$.

The filter's live estimate is advanced to the required horizon by advancing the phase state directly and re-evaluating the harmonic sum at that advanced phase:
\vspace{-2mm}
\begin{align}
\theta &\rightarrow \theta + \omega_r \tau_{horizon}, \\ 
r(t) &= \hat{x}(t+\tau_{horizon}) \nonumber \\
&= a_0 + \sum_{k=1}^{K} A_k \sin\!\big(k(\theta + \omega_r \tau_{horizon}) + \phi_k\big)
\label{eq:prediction}
\end{align}
The measurement depends on $\theta$ and $\phi_1$ only through their sum, $\theta + \phi_1$ (and likewise $k\theta+\phi_k$ for every harmonic). If $\theta$ advances at an incorrect rate due to an erroneous $\omega_r$, the filter can absorb the resulting error by letting $\phi_1$ drift to compensate, and the fitted measurement $h(\mathbf{s})$ still matches the sensor closely. This failure is invisible in the residual and in the tracked signal itself; it only appears in $\omega_r$'s own trajectory, or in the forecast, which advances $\theta$ by $\omega_r \tau_{horizon}$ alone and gets none of $\phi_1$'s compensation. A filter can therefore look like it is tracking correctly while forecasting at an intirely incorrect rate, with nothing in the fit quality to indicate it.

To prevent this, $\omega_r$ is constrained after each update step to stay within a fixed fraction $\beta$ of the rate $\hat\omega_r$,
\begin{equation}
\omega_r \in \big[\hat\omega_r(1-\beta),\ \hat\omega_r(1+\beta)\big]
\label{eq:omega_bounds}
\end{equation}
A fixed absolute range would not work across subjects, since resting breathing rates vary widely between patients. Bounding relative to each subject's own estimate keeps the constraint meaningful regardless of who is being tracked.

A further diagnostic follows directly from quantities already defined above. The normalized innovation squared,
\begin{gather}
    \mathrm{NIS}_k = \frac{\nu_k^2}{S_k}, \label{eq:nis}
\end{gather}
compares each innovation to the variance the filter itself predicted for it (Eqs.~\ref{eq:ekf_innovation}–\ref{eq:ekf_innov_cov}). Because the measurement is scalar, $\mathrm{NIS}_k$ is chi-squared distributed with one degree of freedom under a correctly specified filter, with expectation 1. A sustained rise above that expectation indicates the filter's model no longer matches the observed signal, whether from mistuning or from a change in the signal.

\subsubsection{Compensation and Needle Drive}\label{subsubsec:sys-id}

When the needle is given a position-control reference, the needle actuation axis is identified as a second-order system from its velocity step response. Since position is the time-integral of velocity, the plant seen by the
position-tracking loop carries an additional integrator:
\begin{equation}
G(s) = \frac{K_p\,\omega_n^2}{s\,\left(s^2+2\zeta\omega_n s+\omega_n^2\right)}
\label{eq:plant}
\end{equation}
Lead compensation is applied to reduce the tracking lag of this loop:
\begin{equation}
C(s) = K_c\,\frac{s+z}{s+p}, \qquad p > z
\label{eq:lead_compensator}
\end{equation}
$z$, $p$, and $K_c$ are chosen by a constrained search. 
Candidates are restricted to those achieving at least a target phase margin and a gain-crossover frequency held safely below the plant's natural frequency $\omega_n$. Among the candidates that satisfy both, the one giving the lowest residual closed-loop tracking
lag $\tau_{cl}(\omega_r)$ at the nominal breathing rate is selected, subject to a further
constraint bounding how much the compensator's high-frequency gain may amplify real
position-measurement noise into spurious corrections.

\subsection{Procedural State Machine}
The biopsy procedure proceeds through four sequential steps following the position of the robot in the CT bore~\cite{zhang2020biopsy}: (1) torso approach and contact, (2) breathing cycle sensing, (3) initial drive and dermal puncture, and (4) gated needle drive to the target depth. 

\subsubsection{Torso Approach and Contact}
The Time-of-Flight (ToF) sensor guides the wrist toward the torso at a set velocity. Upon contact, as detected by the contact sensor, the breathing detection informs incremental advancements until 3-4 full breaths are detected.

\subsubsection{Breathing Cycle Sensing}
The contact sensor data feeds the estimator and lead compensator. To ensure tracking stability, advancement to active insertion requires three simultaneous criteria:
\begin{itemize}
\item Convergence: The EKF runs for a minimum number of breath cycles after being seeded before its estimate is trusted.
\item Consistency: The windowed mean Normalized Innovation Squared (NIS) falls within a configured band.
\item Accuracy: The forecast standard deviation at horizon $\tau_{horizon}$ drops below a preset noise budget.
\end{itemize}

\subsubsection{Initial Drive and Dermal Puncture}
The needle punctures the dermal layer during the exhalation basin, stopping prior to inhalation. This mounts the needle to the tissue and prevents needle slippage during breathing motions. The exhalation basin is targeted because patients exhibit significantly higher breath-to-breath consistency at full exhale compared to full inhale.

\subsubsection{Gated Insertion}
Using the predictive controller (Section~\ref{subsec:controls}), the needle drives incrementally during the exhalation basin of each breath until reaching the target depth. The exhalation basin is targeted for the same reason as in the initial drive.

%% file: sections/4.0_results.tex
\section{Experiments and Results}
\subsection{Characterization}
\begin{table}
\vspace{3mm}
\centering
\caption{Sensor Noise}
\label{tab:sensor-noise}
\footnotesize
\begin{tabular}{c|c c}
 Name & Peak-to-Peak (mm) & SD (mm) \\
\hline
ToF Sensor & 8 & 2 \\
Contact Sensor & 0.011 & 0.002 
\vspace{-7mm}
\end{tabular}
\end{table}

\subsubsection{Sensor Characterization}
The Time-of-Flight (ToF) sensor was calibrated for non-linear deviation at 10 stops between 0 mm to 10 mm from the phantom surface, which yielded an offset of 2.6 mm and a slope of 1.05. The contact sensor's hall effect sensor is not affected by distance; therefore no non-linear deviation was measured during testing. The stationary noise of both sensors was measured and recorded in Table~\ref{tab:sensor-noise}. The ToF sensor had peak-to-peak noise up to 8 mm with a standard deviation of 2 mm, supporting its use when traversing large distances. The contact sensor had a peak-to-peak range of 0.011 mm with a standard deviation of 0.002 mm, lower than the 0.37 mm minimum requirement listed in Table~\ref{tab:breathing_summary}.

\subsubsection{Lag Compensation}
Through a normalized cross-correlation between a known commanded signal on the actuator and the corresponding sensor reading, the system's latency was calculated to be approximately 300 ms. This latency accounts for any delay caused by mechanism damping, signal transmission, and signal processing delays. The contact mechanics acount for a majority of this delay with the contact sensor running at 100 Hz. This latency is similarly found on the CT bore robot mechanism to be around 330 ms.

\subsubsection{System identification}
System identification was performed on the needle-driving mechanism to characterize its dynamics under motor actuation. The axis was given a step velocity command from rest, and the resulting transient was fit to a standard second-order model 
as outlined in Sec.~\ref{subsubsec:sys-id}. 10 trials were conducted in free air, yielding measured plant values $K_p = 0.92$, $\omega_n = 27.0$ rad/s, and $\zeta = 0.35$. The trials were repeated while the needle was inserted into the tissue which offers more resistance to the motion. However, no significant difference in system dynamics was found, and the values above were used across all phases of the needle insertion sequence. 

\subsection{Approach Validation}
In order to validate the transition from the Time of Flight sensor to the touch sensor during the Approach to Contact step, a breathing motion was applied to the phantom, and 10 trials were completed across five patient profiles. The final push is completed in increments that detect if the full breath cycle is measured, pushing closer if the touch sensor is fully depressed. In 10/10 trials, the needle was successfully brought from randomized distances between 30 and 40 mm away to a standoff distance of 1 mm with a tolerance of $\pm 0.3$ mm with no tissue damage.

\subsection{EKF Tuning}
\begin{figure}[t]
\vspace{2mm}
    \centering
    \includegraphics[width=0.49\textwidth, trim={0 0 0 0}, clip]{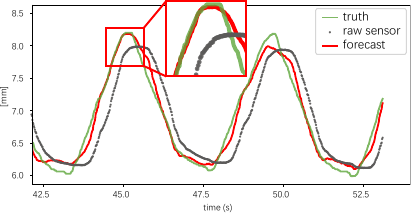}
    \caption{Comparison of the EKF forecast (red) and raw sensor values (gray) overlaid with the ground truth breathing motion (green) shows that the EKF forecast successfully adjusts for lag in sensor readings and closely tracks the ground truth.}
    \label{fig:rmse-comparison}
    \vspace{-5mm}
\end{figure}

\begin{figure*}[t]
\vspace{2mm}
    \centering
    \includegraphics[width=0.99\linewidth]{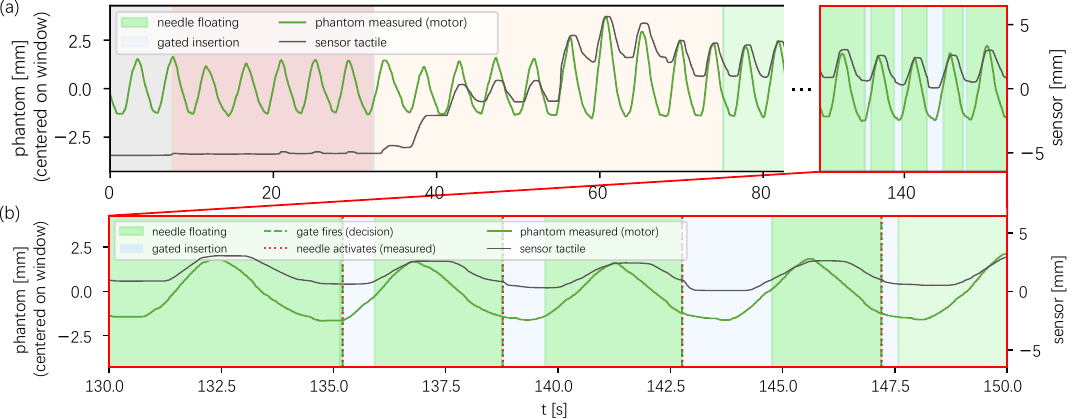}
    \caption{The torso phantom and the measured breathing signal are shown in green and gray above respectively. In (a) the background colors indicate the phase of the sequence. Gray indicates initialization. Red indicates the needle carriage approaching. Orange indicates contact incremental pushes to reach full breath detection. Green indicates full breath detection and EKF settling. Blue slices indicate motions of gated insertion. The gated insertion segment is further detailed in (b), showing insertion highlighted by blue during full exhales.}
    \label{fig:breathing-results}
    \vspace{-5mm}
\end{figure*}

The process noise covariance matrix and velocity bounds written in~\eqref{eq:Q_construction} and~\eqref{eq:omega_bounds} were calibrated by adjusting $q_{scale}$ and $\beta$, respectively, with 10 trials across five different patient profiles, sweeping the process noise scale between 0.3 and 1.0 and the velocity bounds between 0\% and 30\%. The calibrated values of 0.5 and 10\% held frequency lock across all 10 calibration trials, with a mean forecast RMSE of 0.293 mm (max 0.591 mm). 
The tuned parameters were then verified on five different patient profiles, giving a mean forecast RMSE of \textbf{0.165 mm compared to 0.235 mm} for the raw sensor reading. \textbf{The EKF forecast resulted in a 29.8\% reduction in error and outperformed the raw sensor input on 10/10 validation profiles.}

As seen in Figure~\ref{fig:rmse-comparison}, the EKF forecast (red) closely tracked the ground truth waveform (green), made more significant when considering the raw sensor values (gray). The EKF forecast also showed consistent predictions across multiple breathing cycles.

\subsection{Needle Insertion}
\subsubsection{Insertion Detection Sensitivity}
The needle was driven through the phantom using the needle driving motor while holding the carriage and breathing motor stationary. The torque was recorded at three different angles and two different speeds for a total of six trials.
During insertion, the torque was observed to increase in proportion to the depth of insertion as the needle pushed into the phantom. After puncturing the back of the phantom, oscillating drops in torque were measured as the needle dragged and slipped through the phantom. This torque measurement has potential for informing the controls algorithm, but is currently limited by motor hardware for deployment on the CT bore robot mechanism. 

\subsubsection{Full Sequence}
The full sequence was tested end to end on five trials with an initial sinusoid signal and then patient profiles driving to a total depth of 40~mm in two stages: an initial drive, followed by a series of gated insertions continuing until the full depth was reached. Both stages used the same firing gate, requiring the forecast to land within the bottom 15\% of the tracked breathing excursion with a forecast standard deviation under 0.5~mm before driving, and the same model-free abort rule during a drive.

The initial drive inserted an average of 17.91 mm across five trials, ensuring that the needle punctured through the torso phantom to keep the needle seated in the tissue firmly.
From there, one trial required two more gated insertions to reach the required depth while the rest took three more insertions. All five trials successfully reached the target depth while not violating the gating parameters. The average time till full insertion is 14.41s.

The needle driving mechanism was commanded to be compliant between gated insertions as to float along with the breathing motion. This was observed to be more consistent during exhalation compared to inhalation. We identified the root cause to come from an unrealistic torso phantom that deformed greatly and a needle driving mechanism that offered high resistance. 

\subsection{Anomaly Detection}
To verify the sensing system detects unsafe deviations from normal breathing, four synthetic disturbances: an amplitude step (2.0$\to$3.0 mm), a rate step (15$\to$20 bpm), both together, and a bounded rate wander ($\pm4$ bpm about 15 bpm, 8s decorrelation), were introduced into an otherwise quiet baseline waveform at a zero crossing so its position stayed continuous through the change.

The detection algorithm thresholds each run's own $\mathrm{NIS}_k$ (Eq.~\ref{eq:nis}) at the 99.9th percentile of a 1-second sliding mean, learned from that run's clean baseline. Detection triggers on the first sustained 0.2s crossing, timestamped to when the crossing began.
\begin{table}[t]
    \centering
    \begin{tabular}{l|l}
        \textbf{State} & \textbf{Response} \\
        \hline
        Approach & Stop \\
        \hline
        Breath Sensing & Stop \\
        \hline
        Insertion & Slack \\
        \hline
        Breath following (passive) & Do nothing (slack) \\
        \hline
        Gated Insertion & Slack \\
    \end{tabular}
    \caption{State and Abnormal Event Response}
    \label{tab:abnormal-states}
    \vspace{-5mm}
\end{table}

Latency splits into sensing lag (the disturbance reaching the tactile sensor, measured per run) and detector lag (the remainder). The three step disturbances were detected in 0.17-0.23s, matching the sensing lag alone (0.22-0.25s). The delay is calculated through cross correlating the sensor signal to the baseline to account for the mechanical settling time of the phantom skin. When the step is an immediate, large mismatch, the smoothed NIS clears the threshold as soon as it reaches the sensor, resulting in a calculation that can be smaller or larger than the previously calculated sensing lag by a few milliseconds. The rate wander was detected after 2.32s, roughly 2.1s of which is real detector lag, since a slow drift accumulates mismatch gradually rather than all at once. This is by design as rate variability of this size occurs in normal breathing, so responding slowly to a gradual drift and quickly to an abrupt one is intended. NIS peaked 28-562 times over threshold and reconstruction error grew 3.8-19.8 times across all four.

When the breathing detected deviates from the predicted values from the Extended Kalman filter during quiet breathing, the system responds safely following the cases in Table~\ref{tab:abnormal-states}. 

%% file: sections/5.0_discussion_conclusion.tex
\section{Conclusion and Future Work}
In this work, we presented a respiratory-gated needle insertion system designed for integration with a CT bore robot. Our mechanism was first implemented on a testbed with a torso skin phantom to replicate chest movement during quiet breathing, and a needle insertion mechanism with four phases: Approach, Sensing, Initial Drive, and Gated Drive. Our results demonstrated successful needle insertion that satisfied respiratory gating constraints derived from sensed breathing motions. 

Improvements to the testbed systems can be explored in future work, including a more realistic torso phantom that exhibits better skin deformation properties, additional modes of breath sensing, and a lighter needle insertion mechanism. Prior to clinical deployment, we suggest performing patient- and tool-specific system identification at the start of the procedure. For subsequent work on better integration between the CT robot and the needle insertion mechanism, we plan to explore methods of respiratory-compensation subject to an RCM constraint around the insertion site, allowing for insertions at more extreme angles.

\section*{Acknowledgment}

In this paper, Claude Code was utilized to implement the described process and scripts to graph or visualize data. 
Gemini was used for blemish removal in figures.